\def\year{2019}\relax
\documentclass[letterpaper]{article} 
\usepackage{aaai19}  
\usepackage{times}  
\usepackage{helvet}  
\usepackage{courier}  
\usepackage{url}  
\usepackage{graphicx}  

\usepackage{subfigure}
\usepackage{amsmath}
\usepackage{multirow}
\usepackage{mathrsfs}

\begin{document}
%
\title{Read, Watch, and Move: Reinforcement Learning for Temporally Grounding Natural Language Descriptions in Videos}
\author{
Dongliang He$^{\dag}$,
Xiang Zhao$^{\dag}$,
Jizhou Huang$^{\ddag,\dag}$,
Fu Li$^{\dag}$,
Xiao Liu$^{\dag}$,
Shilei Wen$^{\dag}$
\\
$^{\dag}$Baidu Inc., Beijing, China \\
$^{\ddag}$Research Center for Social Computing and Information Retrieval, Harbin Institute of Technology, China \\
\{hedongliang01, zhaoxiang05, huangjizhou01, lifu, liuxiao12, wenshilei\}@baidu.com
}

\maketitle
\begin{abstract}
The task of video grounding, which temporally localizes a natural language description in a video, plays an important role in understanding videos.
Existing studies have adopted strategies of sliding window over the entire video or exhaustively ranking all possible clip-sentence pairs in a pre-segmented video, which inevitably suffer from exhaustively enumerated candidates.
To alleviate this problem, we formulate this task as a problem of sequential decision making by learning an agent which regulates the temporal grounding boundaries progressively based on its policy.
Specifically, we propose a reinforcement learning based framework improved by multi-task learning and it shows steady performance gains by considering additional supervised boundary information during training.
Our proposed framework achieves state-of-the-art performance on ActivityNet'18 DenseCaption dataset \cite{densecaption} and Charades-STA dataset \cite{charades,tall} while observing only 10 or less clips per video.
\end{abstract}

\section{Introduction}
The task of grounding natural language in image/video, which aims to localize a short phrase \cite{rohrbach2016grounding,endo2017attention,tellex2009towards,regneri2013grounding} or a referring expression \cite{bojanowski2015weakly,gao2016acd} in a given image/video, is essential for understanding the way human perceive images and scenes. In the past few years, studies on detecting a particular object in image \cite{Pirinen_2018_CVPR} or activity in video \cite{lifeifei_loc,localization_multi_stream,ssn,cdc} have been extensively exploited. However, they are restricted to a predefined close set of object/activity space. In this paper, we take a further step by focusing on the problem of temporally grounding an open-ended natural language description in a long, untrimmed video.

\begin{figure}[!t]
\centering
\includegraphics[width=\columnwidth]{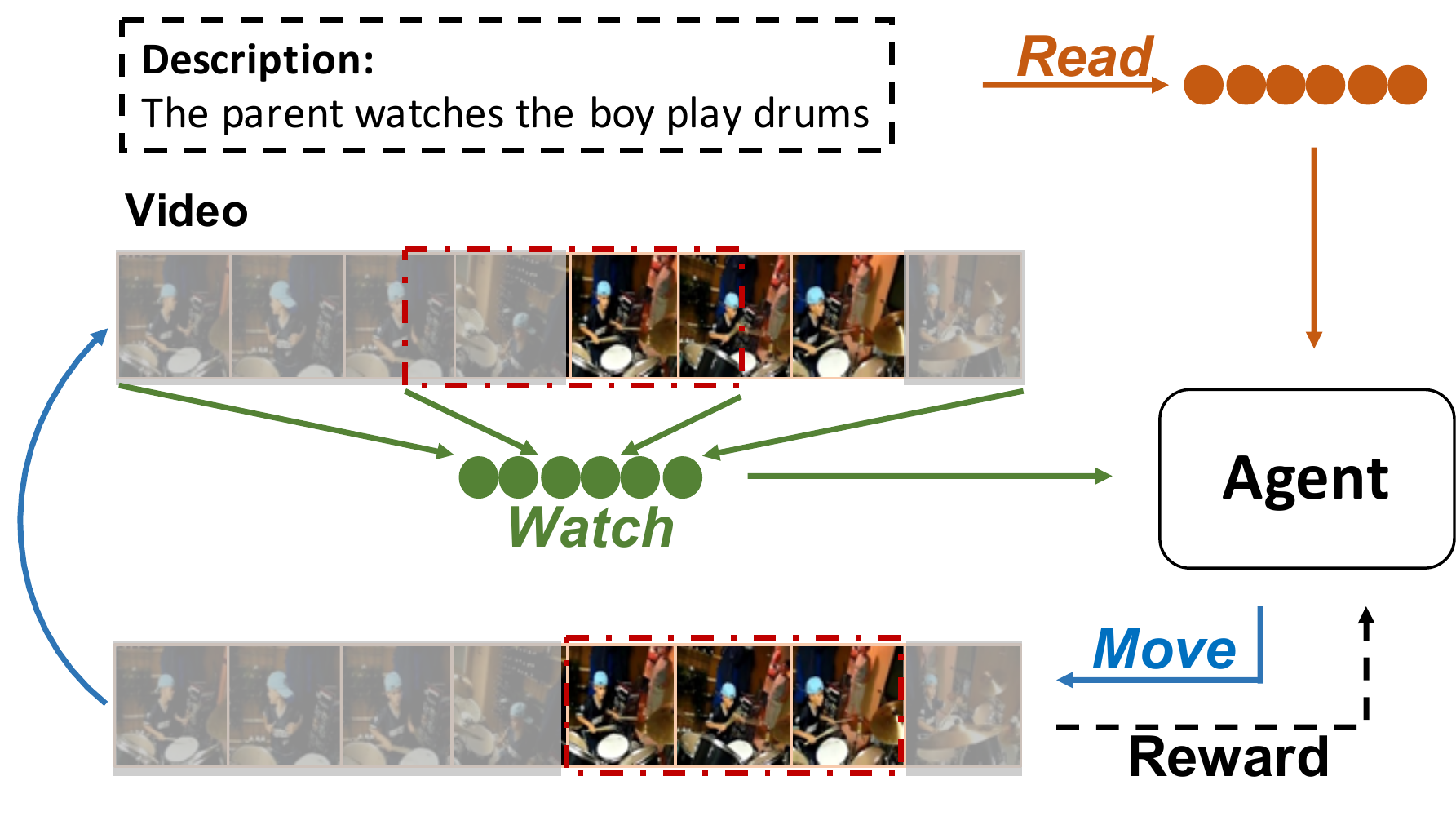}
\caption{Illustration of the proposed formulation of read, watch, and move with reinforcement learning.
At each time step, the agent makes an observation, takes an action, and receives a reward evaluating that action. The policy is updated according to this reward.}
\label{fig:example}
\end{figure}

Specifically, given an untrimmed video and a natural language description, our task is to determine the start and end time stamps of a clip in this video that corresponds to the given description. Exploration on natural language enables us to deal with not only an open set of objects/activities, but also the correlations and interactions between the involved entities. This is a challenging task as it requires both vision and language understanding, but it has important applications in video understanding tasks such as video retrieval and text-oriented video highlight detection. Despite the fact that both action localization and video grounding tasks share the same goal of extracting a well-matched video clip w.r.t. a given query, the methods for action localization are not readily applicable to video grounding. The main reason is that it requires to map natural language descriptions to a close set of labels before applying these methods, which inevitably causes information loss. For example, if the sentence ``The parent watches the boy play drums'' is mapped to an action label ``play drums'', action detectors can hardly tell the start point of the presence of the parent.

Although there is a growing interest in grounding natural language in videos, e.g., \cite{tall,mcn,retrieval_attentive}, these methods have a limitation that they can be applied only when all candidates are exhaustively enumerated, as they need to slide over the entire video or rank all possible clip-sentence pairs to be successful in generating the final grounding result.
Moreover, ranking strategy \cite{mcn} is specially designed for pre-segmented videos and is not applicable to general cases. To effectively detect the temporal boundaries of a given description from a video, especially from a long, untrimmed video, it is important to make judgments only by several glimpses. To the best of our knowledge, the problem of grounding natural language in videos without sliding over the entire video has not been well addressed.

In this paper, we propose an end-to-end reinforcement learning (RL) based framework for grounding natural language descriptions in videos.
As shown in Figure \ref{fig:example}, the agent iteratively reads the description and watches the entire video as well as the temporary grounding clip and its boundaries, and then it determines where to move the temporal grounding boundaries based on the policy. At each time step, the agent makes an observation that details the currently selected video clip, and takes an action according to its policy. The environment is updated accordingly and offers the agent a reward that represents how well the chosen action performs.
Our key intuition is that given a description and the current grounding start and end points, human can make hypotheses on how these points should be moved after observing a clip, and then iteratively adjust the grounding boundaries step by step towards finding the best matching clip. Therefore, we naturally formulate this task as learning an agent for sequential decision making to regulate the temporal boundaries progressively based on the learned policy. In addition, we introduce two distinctive characteristics into the RL framework. First, we improve the framework with multi-task learning and show steady performance gains by considering additional supervised boundary information during training. Second, we introduce a location feature, which enables the agent to explicitly access where the current grounding boundaries are.

Specifically, we design an observation network that takes visual and textual information as well as current temporal boundaries into consideration. Given the observation, the decision making policy is modeled by a policy network that outputs the probability distribution over all possible actions, and then the agent adjusts the grounding boundaries by taking an action according to the policy.
The policy network is implemented using a recurrent neural network (RNN).
In addition, in order to achieve better feature learning, we propose to leverage the ground truth for supervised temporal IoU (tIoU) regression as well as boundary regression at each step. Our RL model can be trained end-to-end for natural language description grounding within several steps of glimpses (observations) and temporal boundaries adjustments (actions). We evaluate our model on two well-known datasets, i.e., ActivityNet DenseCaption \cite{densecaption} and Charades-STA \cite{charades,tall}. Experimental results show that our method achieves state-of-the-art performance on these datasets and validate the superiority of our method.

Here we summarize our main contributions:
\begin{itemize}
\item We propose a reinforcement learning based framework for the task of grounding natural language descriptions in videos.
\item Supervised learning is combined with reinforcement learning in a multi-task learning framework, which helps the agent obtain more accurate information about the environment and better explore the state space to encounter more reward, thus it can act more effectively towards task accomplishment.
\item Experiments conducted on two well-known datasets demonstrate that the proposed method significantly outperforms state-of-the-art method by a substantial margin, which verifies the effectiveness of our method.
\end{itemize}

\begin{figure*}[t]
\begin{center}
\includegraphics[width=\linewidth]{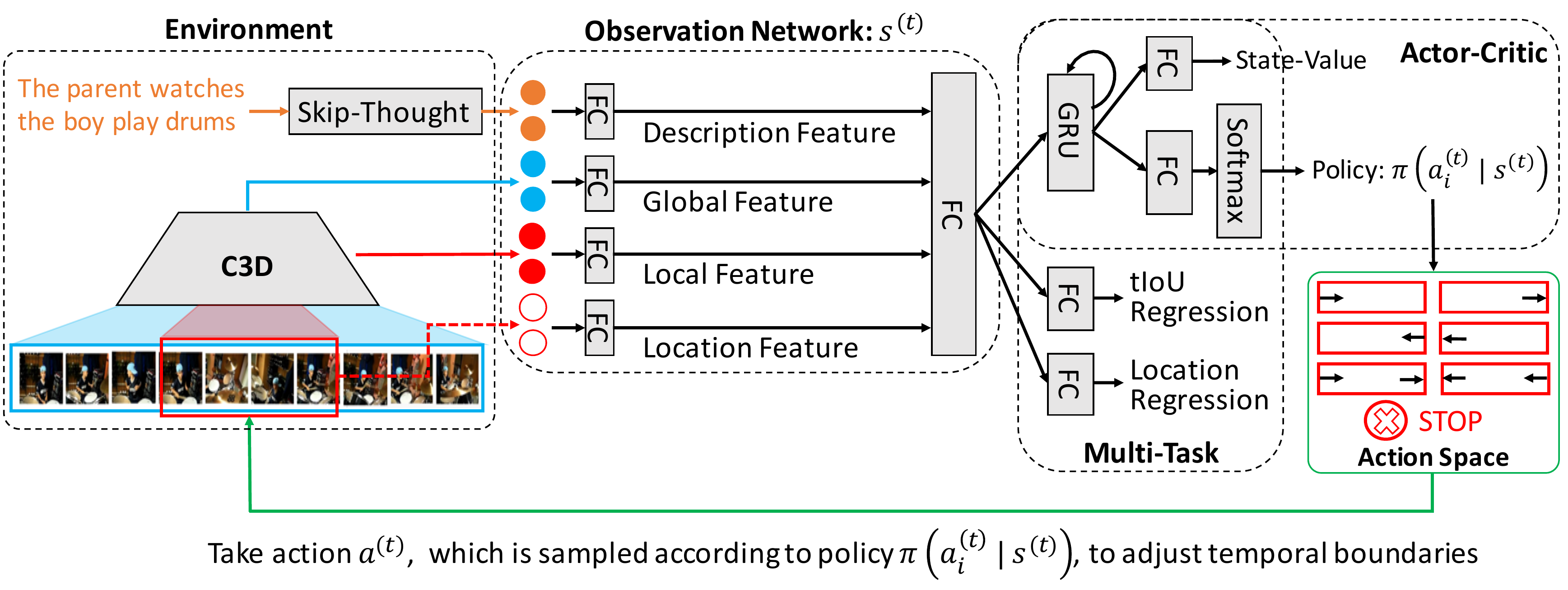}
\end{center}
   \caption{Architecture of the RL-based framework. The action space consists of 7 different ways to adjust the temporal boundaries. State vector $s^{(t)}$, which fuses description embedding feature, global video feature, local video feature, and current location embedding feature, is fed into the actor-critic \cite{li2017deep} to learn both a policy and a state-value. Temporal IoU between current video clip and the ground truth is regressed with supervised learning, so are the temporal localization boundaries.}
\label{fig:overview}
\end{figure*}

\section{Related Work}
We first introduce the related work of grounding natural language in visual content. Then, we introduce the related work on applying reinforcement learning for vision tasks.

The problem of grounding natural language descriptions in videos has attracted an increasing interest more recently. Previous studies typically treat it as a regression or ranking problem by adopting strategies of sliding window over the entire video \cite{tall,retrieval_attentive} or exhaustively ranking all possible clip-sentence pairs in a pre-segmented video \cite{mcn}. However, it is difficult to attain a reasonable compromise between accuracy and computational efficiency due to they inevitably suffer from exhaustively enumerated candidates and the false positive problem. To alleviate this problem, we formulate the task as a sequential decision making problem, rather than treat it as a regression or ranking problem. To this end, we propose an effective method to address this problem with a reinforcement learning based framework.

Our work is also related to the work on understanding visual content. Here we briefly review the closely related work including image grounding, image/video captioning, video retrieval, and temporal action localization in videos.
The task of image grounding, which aims to localize a short phrase or an expression in an image, is a well-studied field, e.g., \cite{plummer2015flickr30k,rohrbach2016grounding,Xiao_2017_CVPR,endo2017attention}. However, they have a different focus from ours.
The task of image/video captioning, which aims to generate a description in text from visual data, has been well studied in computer vision, e.g., \cite{xu2015show,pan2016jointly,gan2017semantic,Liang_2017_ICCV}. The task of visual grounding can be regarded as the conjugate task of visual description generation.
Sentence based video retrieval methods \cite{xu2015jointly,lin2014visual,bojanowski2015weakly} use skip-thought \cite{skipthought} or bidirectional LSTM to encode text queries, and incorporate video-language embedding to find the video that is most relevant to the query. They mainly focus on retrieving an entire video for a given query, rather than finding a temporal segment that matches a given description from a video.
The task of temporal action localization is defined as retrieving video segments relevant to a given action from a predefined close set of action labels \cite{localization_multi_stream,SSAD,ssn,lifeifei_loc}. There is also some work on retrieving temporal segments relevant to a sentence within a video in constrained settings. For example, \citeauthor{tellex2009towards}~\shortcite{tellex2009towards} studied the problem of retrieving video clips from home surveillance videos using text queries from a fixed set of spatial prepositions. \citeauthor{bojanowski2015weakly}~\shortcite{bojanowski2015weakly} proposed to align video segments with a set of sentences which is temporally ordered. Our work is different in that we temporally ground open-ended, arbitrary natural language descriptions in videos.

The use of reinforcement learning \cite{williams1992simple} in our task is inspired by the recent remarkable success of applying it in various vision tasks. The goal of reinforcement learning is to learn an agent to evaluate certain actions under particular states or learn an action policy with respect to current states. It is effective to optimize the sequential decision making problems. Recently, it has been successfully applied to learn task-specific policies in various computer vision problems. For example, spatial attention policies are learned for image recognition in \cite{mnih2014recurrent} and \cite{liu2017localizing}. RL framework is also leveraged for object detection in \cite{Pirinen_2018_CVPR}. \citeauthor{Li_2018_CVPR}~\shortcite{Li_2018_CVPR} proposed an aesthetic aware reinforcement learning framework to learn the policy for image cropping and achieved promising results. As for vision tasks in video, \cite{lifeifei_loc} proposed to use  RL for action localization in videos and achieved a good trade-off between localization accuracy and computation cost. Recently, video object segmentation with deep reinforcement cutting-agent learning is explored in \cite{han2018reinforcement}, and hierarchical reinforcement learning \cite{li2017deep} is incorporated for video description in \cite{Wang_2018_CVPR}. Our video grounding task can be formulated as a sequential decision making problem, thus it fits naturally into the reinforcement learning paradigm. While conceptually similar, our model is novel in that it successfully combines supervised learning with reinforcement learning in a multi-task learning framework, which helps the agent obtain more accurate information about the environment and better explore the state space to encounter more reward, thus it can act more effectively towards task accomplishment.

\section{Methodology}
Given the insight that human usually localize a description in a video by following the iterative progress of observing a clip, making a hypothesis on how the boundaries should be shifted, and verifying it, we formulate the task of  temporally grounding descriptions in videos as a sequential decision-making problem. Therefore, it naturally fits into the RL framework. Figure \ref{fig:overview} shows the architecture of the proposed model.
The temporal grounding location is initially set to $[l_s^{(0)}, l_e^{(0)}]$. At each time step, the observation network outputs the current state of the environment for the actor-critic module to generate an action policy. The policy defines the probabilistic distribution of all the actions in the action space and can be easily implemented by softmax operation. The agent samples an action according to the policy and adjusts the temporal boundaries iteratively until encountering a STOP action or reaching the maximum number of steps (denoted by $T_{max}$).

\subsection{Environment and Action Space}
In this task, the action space $\mathcal{A}$ consists of 7 predefined actions, namely, \emph{moving start/end point ahead by $\delta$ }, \emph{moving start/end point backward by $\delta$}, \emph{shifting both start and end point backward/forward by $\delta$}, and a \emph{STOP} action. In this paper, $\delta$ is set to $N/10$, where $N$ is the video length. In fact, the two shifting actions can be accomplished by two successive moving actions, and we define them mainly for the purpose of enabling a faster temporal boundary adjustment.

Once an action is executed, the environment changes accordingly.
Environment here is referred to as the combination of the video, the description, and the current temporal grounding boundaries. We use skip-thought \cite{skipthought} to generate the sentence embedding $E$ and feed the entire video into C3D \cite{c3d} to obtain the global feature vector $V_G$. At each step, these features are retained. To represent the status of the current temporal grounding boundaries for the $t^{th}$ step of decision, we propose to explicitly involve the normalized start point and end point $L^{(t-1)}=[l_s^{(t-1)}, l_e^{(t-1)}], t=1,2,...,T_{max}$ as supplementary to the local C3D feature $V_L^{(t-1)}$ of the current video clip. In the observation network, sentence embedding, global video feature, local video feature, and normalized temporal boundaries are encoded using fully-connected (FC) layers and their outputs are concatenated and fused by a FC layer to generate the state vector $s^{(t)}$, which is calculated as:
\begin{equation}
s^{(t)} = \Phi(E,V_G,V_L^{(t-1)},L^{(t-1)}).
\end{equation}

\subsection{Advantage Actor-Critic RL Algorithm}
It has been proven that actor-critic algorithm \cite{li2017deep} is able to reduce the variance of the policy gradient. As shown in Figure \ref{fig:overview}, the sequential decision making process is modeled by an actor-critic module, which employs GRU \cite{cho2014learning} and FC layers to output the policy $\pi(a_i^{(t)}|s^{(t)}, {\theta_\pi})$ and a state-value $v(s^{(t)} | \theta_v)$. The policy determines the probabilistic distribution of actions and the state-value serves as estimation of the value of the current state (i.e., critic). $\theta_\pi$ denotes parameters of the observation network and the policy branch, $\theta_v$ denotes parameters of the critic branch. The input of the GRU at each time step is the observed state vector $s^{(t)}$. The hidden state of the GRU unit is used for policy generation and state-value estimation.

\subsubsection{Reward}
Our goal is to learn a policy that can correctly grounding descriptions in videos, so the reward for the reinforcement learning algorithm should encourage the agent to find a better matching clip step by step. Intuitively, allowing the agent to search within more steps could yield a more accurate result but it will introduce more computational cost. To alleviate this issue, we reward the agent for taking better actions:
the agent will get a positive reward 1 if the temporal IoU (tIoU) between $L^{(t)}=[l_s^{(t)}, l_e^{(t)}]$ and the ground-truth grounding location $G = [g_s, g_e]$ gets higher after the $t^{th}$ step of action. On the contrary, we punish the agent for taking worse actions: the agent  will get no reward if the tIoU gets lower. To balance the number of actions and the accuracy, we further punish the number of steps by a negative reward of $-\phi * t$, where $\phi$ is the positive penalty factor ranging from 0 to 1. Moreover, there are chances that the start point gets higher than the end point after taking several actions, which is an undesirable state. To avoid such situations, the reward function should punish the agent if the tIoU is a negative value.
To sum up, the reward of the $t^{th}$ step is formulated as:
\begin{equation}
r_t =
\begin{cases}
{1-\phi * t,} & {tIoU^{(t)} > tIoU^{(t-1)}} \geq 0 \\
{-\phi * t}, &  0 \leq {tIoU^{(t)} \leq tIoU^{(t-1)}} \\
{-1-\phi * t,} & otherwise
\end{cases},
\end{equation}
where $tIoU$ is calculated as:
\begin{equation}
tIoU^{(t)} = \frac{min(g_e,l_e^{(t)})-max(g_s,l_s^{(t)})}{max(g_e,l_e^{(t)})-min(g_s,l_s^{(t)})}.
\end{equation}

The task is modeled as a sequential decision making problem, whose goal is to finally hit the ground truth, so the reward of following up steps should be traced back to the current step. Therefore, our final accumulated reward function is designed as follows:
\begin{equation}
R_t =
\begin{cases}
r_t + \gamma*v(s^{(t)}|\theta_v), & t = T_{max} \\
r_t + \gamma*R_{t+1}, & t=1,2,...,T_{max}-1
\end{cases},
\end{equation}
where $\gamma$ is a constant discount factor.

\subsubsection{Objective}
Instead of directly maximizing the expected reward, we introduce the advantage function \cite{li2017deep}, and our objective becomes maximizing the expected advantage $J_a(\theta)$:
\begin{equation}
J_a(\theta_\pi) = \sum_{{a}\in \mathcal{A}}\left\{\pi(a|s,\theta_\pi)(R-v(s|\theta_v))\right\}.
\end{equation}

The gradient of $J_a$ is:
\begin{equation}
\begin{aligned}
\nabla_{\theta_\pi} & J_a(\theta_\pi) = \\
& \sum_{a \in \mathcal{A}}{\pi(a|s,\theta_\pi)\left(\nabla_{\theta_\pi}{\log\pi(a|s,\theta_\pi)}\right)(R-v(s|\theta_v))}.
\end{aligned}
\end{equation}

It is difficult to directly optimize this problem due to the high dimension of the action sequence space. In RL, Mente Carlo sampling is adopted and the policy gradient is approximated as:
\begin{equation}
\nabla_{\theta_\pi} J_a(\theta_\pi) \approx {\sum_{t}\nabla_{\theta_\pi}{\log{\pi(a_i^{(t)}|s^{(t)},\theta_\pi)}}(R_t-v(s^{(t)}|\theta_v))}.
\end{equation}

Therefore, the maximizing problem can be optimized by minimizing the following loss function $\mathcal{L}_{A}'$ using stochastic gradient descent (SGD):
\begin{equation}
\label{eq:obj1}
\mathcal{L}_{A}'(\theta_\pi) = -{\sum_{t}{\left(\log{\pi(a_i^{(t)}|s^{(t)},\theta_\pi)}\right)}(R_t-v(s^{(t)}|\theta_v))}.
\end{equation}

We also follow the practice of introducing entropy of the policy output $H(\pi(a_i^{(t)}|s^{(t)},\theta_\pi))$ into the objective, since maximizing the entropy can increase the diversity of actions \cite{Li_2018_CVPR}. Therefore, the overall loss of the actor branch becomes:
\begin{equation}
\mathcal{L}_{A}(\theta_\pi) = \mathcal{L}_{A}'(\theta_\pi) - \sum_{t}{{\lambda}_0 * H(\pi(a_i^{(t)}|s^{(t)},\theta_\pi))},
\end{equation}
where $\lambda_0$ is a constant scaling factor and it is set to 0.1 in our experiments.

In the context of actor-critic algorithm, the objective of the estimated state-value $v(s^{(t)}|\theta_v)$ by critic branch is to minimize the mean squared error (MSE). Therefore, we introduce the MSE loss for the critic branch defined as follows:
\begin{equation}
\label{eq:loss_mse}
\mathcal{L}_{C}(\theta_v) = \sum_{t}{\left(v(s^{(t)}|\theta_v)-R_t\right)^2}.
\end{equation}

Finally, the overall loss of the RL algorithm is a combination of the two losses:
\begin{equation}
Loss_1 = \mathcal{L}_{A}(\theta_\pi) + {\lambda}_1 * \mathcal{L}_{C}(\theta_v).
\end{equation}

\subsection{Combination of Supervised Learning and RL}
To learn more representative state vectors, we propose to combine reinforcement learning and supervised learning into a multi-task learning framework. The overall loss for supervised learning is a combination of the tIoU regression loss and the location regression loss:
\begin{equation}
Loss_2 = Loss_{tIoU} + {\lambda}_2 * Loss_{Loc}.
\end{equation}

Here we define the tIoU regression loss as the L1 distance between the predicted tIoU and the ground-truth tIoU, i.e.,
\begin{equation}
Loss_{tIoU} = \sum_{t}|tIoU^{(t-1)} - P_{tIoU}^{(t)}|,
\end{equation}
where $P_{IoU}^{(t)}$ is the predicted tIoU at the $t^{th}$ step. Similarly, the L1 location regression loss can be calculated as:
\begin{equation}
\label{eq:loc_loss}
Loss_{Loc} = \sum_{t}y^{(t)}\left(|g_s - P_{s}^{(t)}|+|g_e-P_e^{(t)}|\right)/2,
\end{equation}
where $P_s^{(t)}$ and $P_e^{(t)}$ are the predicted start and end, respectively. $y^{(t)}$ is an indicator which equals 1 when $tIoU^{(t-1)} > 0.4$ and 0 otherwise.

The proposed model can be trained end-to-end by minimizing the overall multi-task loss $Loss_1 + \lambda_3 * Loss_2$. In our experiments, $\lambda_1$, $\lambda_2$, and $\lambda_3$ are empirically set to 1. The supervised loss enables the algorithm to learn to predict how well the current state matches the ground truth and where the temporal boundaries should be regressed while the RL loss enables it to learn a policy for pushing the temporal boundaries to approach the ground truth step by step. In the training phase, parameters are updated with both supervised gradient and policy gradient. Therefore, the generalization of the trained network can be improved by sharing the representations.

\section{Experiments}
\subsection{Datasets and Evaluation Metric}
The models are trained and evaluated on two well-known datasets, i.e., ActivityNet DenseCaption \cite{densecaption} and Charades-STA \cite{charades,tall}. The DenseCaption dataset contains 37,421 and 17,505 natural language descriptions with their corresponding ground-truth temporal boundaries for training and test, respectively.
The Charades-STA dataset is originally collected for video classification and video captioning. \citeauthor{tall}~\shortcite{tall} processed the dataset to make it suitable for video grounding task. In this dataset, there are 13,898 description-clip pairs in the training set, and 4,233 description-clip pairs in the test set.

Following previous work \cite{lifeifei_loc,tall}, we also evaluate the models according to the grounding accuracy $Acc@0.5$, which indicates whether the tIoU between the grounding result generated by a model and the ground truth is higher than 0.5.

\subsection{Implementation Details}
In the training phase, Adam with initial learning rate of 1e-3 is used for optimization. In the test phase, an action is greedily sampled at each time step: $a^{(t)} = \arg\max_{a_i^{(t)}}\pi(a_i^{(t)}|s^{(t)}, \theta)$. The initial temporal grounding location is set to the central half of the entire video, namely, $L^{(0)} = [N/4, 3N/4]$, where $N$ is the total length of the video. Because no optical flow frames are provided in the DenseCaption dataset, we use only the RGB frames to train our models. However, Charades-STA does not have such limitation, thus all the models trained on it use both optical flow and RGB modalities.

The parameters used in the experiments are set as follows. $T_{max}$ is set to 10. The natural language description is embedded into a 2400-dimension vector via skip-thought \cite{skipthought}. In the observation network, the output size of the FC layer for encoding the description feature is set to 1024. Both global feature and local feature are encoded into 512-dimension vectors by FC layers. The normalized temporal boundary $L^{(t-1)}$ is embedded into a 128-dimension vector with a FC layer. The dimension of the state vector $s^{(t)}$ at the $t^{th}$ step is 1024. The GRU cell which is used to model the sequential decision process gets hidden size of 1024. $\phi$ is set to 0.001. $\gamma$ is set to 0.3 on DenseCaption and 0.4 on Charades-STA.

\begin{table}[t!]
\centering
\begin{tabular}{ c |c |c}
  \hline		
  \multirow{2}{*}{Method} & DenseCaption & Charades-STA \\
  \cline{2-3}
  & $Acc@0.5$ & $Acc@0.5$ \\
  \hline
  \hline
  RANDOM &18.1\% &14.0\%  \\
  FIXED &25.4\% &23.1\% \\
  TALL (\citeauthor{tall}) &26.9\% &25.8\% \\
  MCN (\citeauthor{mcn}) &27.4\% &30.8\% \\
  RL-Loc (\citeauthor{lifeifei_loc}) &34.7\% & 32.5\% \\
  \hline
  Ours  &\textbf{36.9\%} &\textbf{36.7\%} \\
  \hline
\end{tabular}
\caption{The evaluation results of different methods.}
\label{t:compare}
\end{table}

\subsection{Comparison with Baselines}
Action localization methods can be adapted to this task by feeding video and sentence features and then making 0 vs. 1 action localization. In fact, both \citeauthor{tall}~\shortcite{tall} (denoted by TALL) and \citeauthor{mcn}~\shortcite{mcn} (denoted by MCN) have adopted this idea.
We use both methods as baselines and implement them using the source codes released by their authors.
\citeauthor{lifeifei_loc}~\shortcite{lifeifei_loc} (denoted by RL-Loc) uses RL for action localization, we also adapt it for video grounding and use it as a baseline. In addition, to better evaluate our proposed model, we also use two hand-crafted approaches as baselines. First, a random strategy (denoted by RANDOM) is employed by randomly selecting a video clip with half of the entire video length for arbitrary descriptions. Second, a fixed strategy (denoted by FIXED) is employed by selecting the central part with a fixed length ($N/2$ for DenseCaption and $N/3$ for Charades-STA) for any descriptions.

Table \ref{t:compare} shows the evaluation results. We observe that: 1) our method achieves the best performance in comparison with all baselines on both datasets; 2) our method significantly outperforms previous state-of-the-art methods by substantial margins: the accuracy is improved by 10.0\% and 10.9\% over TALL; 9.5\% and 5.9\% over MCN; 2.2\% and 4.2\% over RL-Loc on DenseCaption and Charades-STA, respectively.
The evaluation results demonstrate that our model achieves a significant improvement over all previous state-of-the-art models.
This suggests that it is a more natural formulation of the task by adopting the read, watch, and move strategy with sequential decision making.
As for the inference computation cost, both our model and RL-Loc only need to observe up to 10 clips per video. MCN splits a video into 6 segments and needs to enumerate all 21 possible video clips. TALL adopts sliding windows of 128 and 256 frames with an overlap ratio of 0.8 and needs to examine 45.8 candidates per video on average.


\begin{table}[!t]
\centering
\begin{tabular}{ c | c | c |c |c}
  \hline	
  \multirow{2}{*}{$\phi$} & \multicolumn{2}{c|}{DenseCaption} & \multicolumn{2}{c}{Charades-STA}\\	
  \cline{2-5}
  & \# Steps & $Acc@0.5$ & \# Steps & $Acc@0.5$ \\
  \hline\hline	
  1 & 1.00&25.4\% &1.0 &17.9\% \\
  0.1 &5.30 &36.1\% &5.17 &36.3\% \\
  0.01 &9.96 &36.7\% &9.97 & \textbf{36.7\%} \\
  0.001 &9.99 & \textbf{36.9\%} &9.99 & \textbf{36.7\%} \\
  \hline
\end{tabular}
\caption{$Acc@0.5$ and average number of glimpses evaluated with different $\phi$ settings.}
\label{t:phi}
\end{table}

\subsection{Ablation Study}
In this section, we perform a series of ablation studies to understand the importance of different factors, reward, and components. To this end, we train multiple variations of our model with different settings or configurations. In the following experiments, we use RGB feature of the videos on DenseCaption while both RGB and flow features on Charades-STA. Specifically, when two modalities are used, both the global feature $V_G$ and the local feature $V_L^{(t-1)}$ are the concatenation of RGB and optical flow features.

\subsubsection{Impact of Penalty Factors}
In the reward function, we have proposed to punish the agent for taking too many steps to strike a desirable trade-off between grounding accuracy and computation cost. In this experiment, we examine the impact of the parameter $\phi$ while keeping $\gamma$ fixed. Table \ref{t:phi} shows the results of $Acc@0.5$ and the average number of steps the agent takes. From the results, we can clearly observe that the average number of steps gets higher when $\phi$ gets lower on both datasets, and the grounding accuracy increases when $\phi$ decreases. Specially, when the penalty factor was set to 1, the average number of steps is 1, which indicates that the agent takes the STOP action at the first step. On the contrary, if $\phi$ was set to 0.001, the average number of steps is 9.99, which indicates that the agent has fully utilized the possible exploring steps (which is up-bounded by $T_{max}=10$) to adjust the temporal grounding boundaries, thus the best accuracy is achieved.
This shows that $\phi$ can be utilized to strike a desirable trade-off between the accuracy and computation cost. For example, setting $\phi$ to 0.1 could be a balanced choice if we would like to achieve better grounding accuracy in fewer steps.

\begin{figure}[!t]
\begin{center}
\includegraphics[width=0.98\columnwidth]{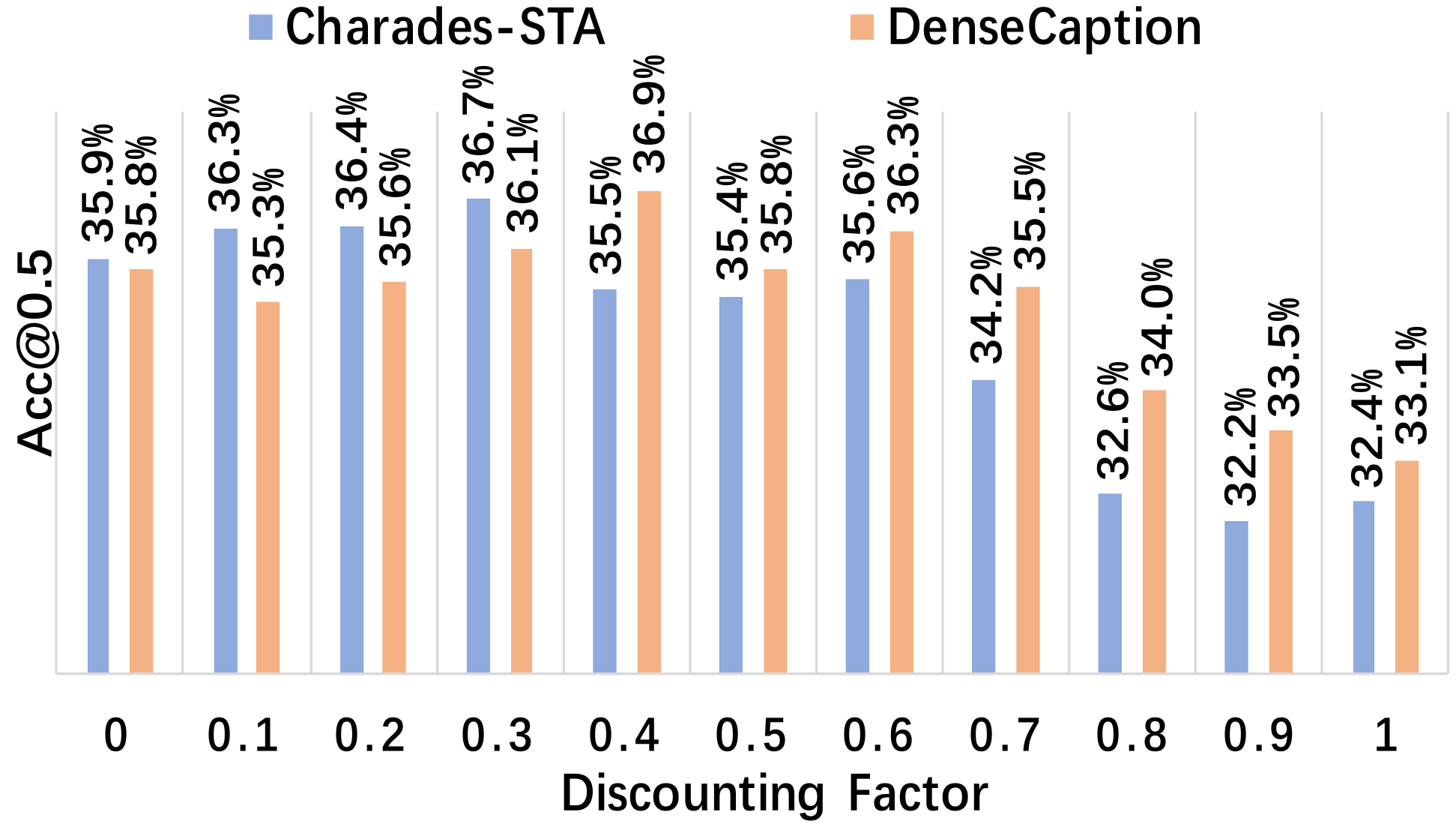}
\end{center}
\caption{Results of the methods with different discount factor $\gamma$ settings on Charades-STA and DenseCaption.}
\label{fig:gamma}
\end{figure}

\subsubsection{Impact of Accumulated Reward}
In order to build connections inside the action sequence, we propose to trace reward of the current step back to the previous steps, so that the reward can be accumulated. In this experiment, we fix $\phi$ to 0.001 and evaluate the impact of the discount factor $\gamma$ in our final reward $R_t$. We range $\gamma$ from 0 to 1 with a step size of 0.1. Then, we train multiple variations of our model with different discounting factors, and examine their impact. Figure \ref{fig:gamma} shows the experimental results.
From the results, it is observed that the grounding accuracy obtained with either 0 or 1 of $\gamma$ is lower than most of those obtained with the values of $\gamma$ in between 0 and 1. Intuitively, when $\gamma$ is set to 0, it indicates that the reward of the current action has no impact on its follow-up actions. This is not a plausible reward strategy for the entire sequential decision making process because the dependency between successive actions is not modeled. Conversely, when $\gamma$ is set to 1, it indicates that the reward of the current action largely depends on its successive actions. This is also not a plausible reward strategy because even if the $t^{th}$ action is an unsuccessful attempt, the accumulated reward $R_t$ can be still positive if there is a chance that any of its successive actions get a positive reward. In this experiment, we empirically observed that when $\gamma$ was set to 0.3 on Charades-STA and 0.4 on DenseCaption, the model achieves the best grounding accuracy on each dataset.

\begin{figure*}[t]
\begin{center}
\includegraphics[width=0.825\linewidth]{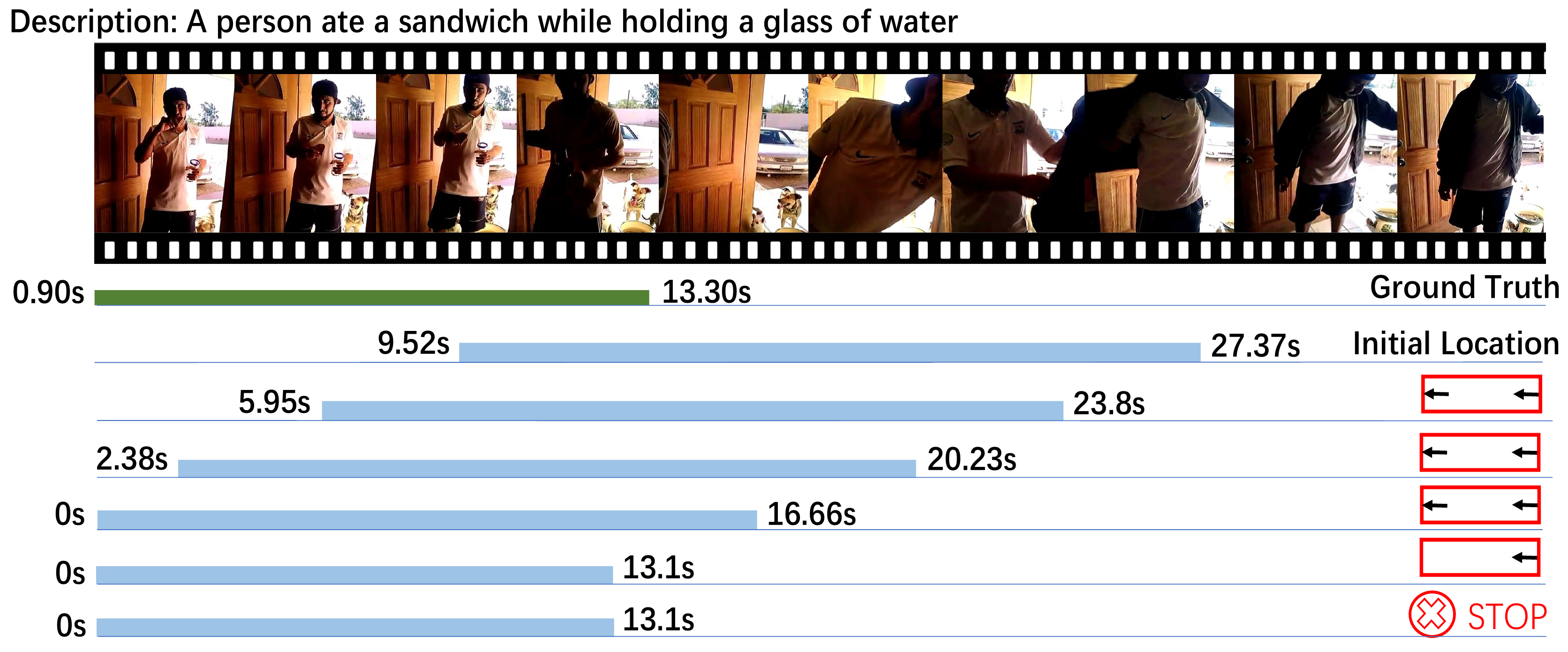}
\end{center}
\caption{Illustration of the action sequence executed when our model grounds the given description in a video.}
\label{fig:demo}
\end{figure*}

\subsubsection{Importance of Different Components}
Here we investigate the effectiveness of each component in our proposed model by conducting an ablation study. In the experiments, $\phi$ was set to 0.001 to enable the agent to fully explore the grounding result within the maximum steps allowed, while $\gamma$ was set to 0.3 on Charades-STA and 0.4 on DenseCaption.

We first study the effectiveness of the introduced location feature. To this end, we train a new model by removing this feature from the observation network, such that only $V_G$, $E$ and $V_L^{(t-1)}$ are encoded at the $t^{th}$ step. Then we investigate the importance of the proposed multi-task learning, which helps the agent to better capture the representation of the current state of the environment. To accomplish this, we train multiple variations of our model by disabling the two modules (tIoU regression and location regression) one after the other while keeping other settings exactly as what they are.
We conduct experiments on both datasets and the results are detailed in Table \ref{t:component}. From the results, we make the following observations.

First, the performance significantly drops on both datasets after disabling all three components, which particularly results in a dramatic decrease of $Acc@0.5$ by 23.7\% (from 36.7\% to 13.0\%) on Charades-STA. The leave-one-out analysis also reveals that the performance drops by a large margin after removing one or more components. This demonstrates that all the three components can significantly contribute to the grounding effectiveness.

Second, we compare the models with or without the location feature used as a component of the environment. The results reveal that the grounding accuracy drops 3.4\% and 1.9\% in terms of $Acc@0.5$ on DenseCaption and Charades-STA, respectively. This verifies that explicit location feature is important to help model the environment for the observation network.
\begin{table}[!b]
\centering
\setlength\tabcolsep{5.5pt} 
\begin{tabular}{ c | c | c |c |c}
  \hline		
  \multicolumn{3}{c|} {Configurations} & \multicolumn{2}{c}{$Acc@0.5$}\\	
  \hline\hline
  Loc & tIoU-R & Loc-R & DenseCaption & Charades-STA \\
  \hline
  $\times$ & $\times$ & $\times$ & 31.5\% & 13.0\% \\
  $\times$ & $\surd$ &$\surd$ &33.5\% &34.8\%\\
  $\surd$ & $\surd$ &$\times$   &35.9\% &35.1\% \\
  $\surd$ & $\times$ & $\times$ &34.5\% &34.9\% \\
  $\surd$ & $\surd$ & $\surd$ &\textbf{36.9\%} &\textbf{36.7\%} \\
  \hline
\end{tabular}
\caption{Results with different network configurations. Loc refers to ``Explicit Location Feature'', tIoU-R refers to ``tIoU Regression'', and Loc-R refers to ``Location Regression''.}
\label{t:component}
\end{table}

Third, the results show that $Acc@0.5$ decreases from 36.9\% to 35.9\% on DenseCaption and from 36.7\% to 35.1\% on Charades-STA after disabling the component of location regression. Moreover, $Acc@0.5$ further reduces to 34.5\% on DenseCaption and 34.9\% on Charades-STA after disabling both the two components of tIoU regression and location regression. This demonstrates that the overall grounding performance will degrade considerably without the consideration of supervised learning. This also confirms the importance of modeling the environment by combining supervised learning with reinforcement learning in a multi-task learning framework, which can help the agent obtain more accurate information about the environment and better explore the state space to encounter more reward.

\subsubsection{Multi-Modality Experiment}
We investigate the effect of combining both RGB and flow features on Charades-STA as a case study. The experimental results show that the grounding accuracies achieved by the models trained with RGB feature and flow feature are 36.2\% and 35.8\%, respectively. However, the grounding accuracy is slightly increased to 36.7\% when the two modalities are fused by feature concatenation. The performance could be further improved with better fusion methods, which we leave as a future work.

\subsection{Qualitative Results}
To visualize how the agent adjusts grounding boundaries step by step, we illustrate an action sequence executed by our agent in the procedure of grounding a natural language description in a video (see Figure \ref{fig:demo}). The agent takes five steps from the initial location and finally grounds the description correctly in the video.

\section{Conclusion}
In this paper, we study the problem of temporally grounding natural language descriptions in videos. This task can be formulated as a sequential decision making process and it  fits naturally into the reinforcement learning paradigm.
Therefore, we model this task as controlling an agent to read the description, to watch the video as well as the current localization, and then to move the temporal grounding boundaries iteratively to find the best matching clip. Combined with supervised learning in a multi-task learning framework, our model can be easily trained end-to-end. Experiments demonstrate that our method achieves a new state-of-the-art performance on two well-known datasets.


\bibliographystyle{./aaai}

\begin{thebibliography}{}

\bibitem[\protect\citeauthoryear{Bojanowski \bgroup et al\mbox.\egroup
  }{2015}]{bojanowski2015weakly}
Bojanowski, P.; Lajugie, R.; Grave, E.; Bach, F.; Laptev, I.; Ponce, J.; and
  Schmid, C.
\newblock 2015.
\newblock Weakly-supervised alignment of video with text.
\newblock In {\em ICCV},  4462--4470.

\bibitem[\protect\citeauthoryear{{Cho} \bgroup et al\mbox.\egroup
  }{2014}]{cho2014learning}
{Cho}, K.; van {Merrienboer}, B.; {Gulcehre}, C.; {Bahdanau}, D.; {Bougares},
  F.; {Schwenk}, H.; {Schwenk}, H.; {Schwenk}, H.; {Bengio}, Y.; {Bengio}, Y.;
  and {Bengio}, Y.
\newblock 2014.
\newblock Learning phrase representations using {RNN} encoder--decoder for
  statistical machine translation.
\newblock In {\em EMNLP},  1724--1734.

\bibitem[\protect\citeauthoryear{Endo \bgroup et al\mbox.\egroup
  }{2017}]{endo2017attention}
Endo, K.; Aono, M.; Nichols, E.; and Funakoshi, K.
\newblock 2017.
\newblock An attention-based regression model for grounding textual phrases in
  images.
\newblock In {\em IJCAI},  3995--4001.

\bibitem[\protect\citeauthoryear{Gan \bgroup et al\mbox.\egroup
  }{2017}]{gan2017semantic}
Gan, Z.; Gan, C.; He, X.; Pu, Y.; Tran, K.; Gao, J.; Carin, L.; and Deng, L.
\newblock 2017.
\newblock Semantic compositional networks for visual captioning.
\newblock In {\em CVPR},  1141--1150.

\bibitem[\protect\citeauthoryear{Gao \bgroup et al\mbox.\egroup }{2017}]{tall}
Gao, J.; Sun, C.; Yang, Z.; and Nevatia, R.
\newblock 2017.
\newblock {TALL}: Temporal activity localization via language query.
\newblock In {\em ICCV},  5267--5275.

\bibitem[\protect\citeauthoryear{Gao, Sun, and Nevatia}{2016}]{gao2016acd}
Gao, J.; Sun, C.; and Nevatia, R.
\newblock 2016.
\newblock {ACD}: Action concept discovery from image-sentence corpora.
\newblock In {\em ICMR},  31--38.

\bibitem[\protect\citeauthoryear{Han \bgroup et al\mbox.\egroup
  }{2018}]{han2018reinforcement}
Han, J.; Yang, L.; Zhang, D.; Chang, X.; and Liang, X.
\newblock 2018.
\newblock Reinforcement cutting-agent learning for video object segmentation.
\newblock In {\em CVPR},  9080--9089.

\bibitem[\protect\citeauthoryear{Hendricks \bgroup et al\mbox.\egroup
  }{2017}]{mcn}
Hendricks, L.~A.; Wang, O.; Shechtman, E.; Sivic, J.; Darrell, T.; and Russell,
  B.
\newblock 2017.
\newblock Localizing moments in video with natural language.
\newblock In {\em ICCV},  5803--5812.

\bibitem[\protect\citeauthoryear{Kiros \bgroup et al\mbox.\egroup
  }{2015}]{skipthought}
Kiros, R.; Zhu, Y.; Salakhutdinov, R.~R.; Zemel, R.; Urtasun, R.; Torralba, A.;
  and Fidler, S.
\newblock 2015.
\newblock Skip-thought vectors.
\newblock In {\em NIPS},  3294--3302.

\bibitem[\protect\citeauthoryear{Krishna \bgroup et al\mbox.\egroup
  }{2017}]{densecaption}
Krishna, R.; Hata, K.; Ren, F.; Fei-Fei, L.; and Niebles, J.~C.
\newblock 2017.
\newblock Dense-captioning events in videos.
\newblock In {\em ICCV},  706--715.

\bibitem[\protect\citeauthoryear{Li \bgroup et al\mbox.\egroup
  }{2018}]{Li_2018_CVPR}
Li, D.; Wu, H.; Zhang, J.; and Huang, K.
\newblock 2018.
\newblock {A2-RL}: Aesthetics aware reinforcement learning for image cropping.
\newblock In {\em CVPR},  8193--8201.

\bibitem[\protect\citeauthoryear{Liang \bgroup et al\mbox.\egroup
  }{2017}]{Liang_2017_ICCV}
Liang, X.; Hu, Z.; Zhang, H.; Gan, C.; and Xing, E.~P.
\newblock 2017.
\newblock Recurrent topic-transition gan for visual paragraph generation.
\newblock In {\em ICCV},  3362--3371.

\bibitem[\protect\citeauthoryear{Lin \bgroup et al\mbox.\egroup
  }{2014}]{lin2014visual}
Lin, D.; Fidler, S.; Kong, C.; and Urtasun, R.
\newblock 2014.
\newblock Visual semantic search: Retrieving videos via complex textual
  queries.
\newblock In {\em CVPR},  2657--2664.

\bibitem[\protect\citeauthoryear{Lin, Zhao, and Shou}{2017}]{SSAD}
Lin, T.; Zhao, X.; and Shou, Z.
\newblock 2017.
\newblock Single shot temporal action detection.
\newblock In {\em ACM MM},  988--996.

\bibitem[\protect\citeauthoryear{Liu \bgroup et al\mbox.\egroup
  }{2017}]{liu2017localizing}
Liu, X.; Wang, J.; Wen, S.; Ding, E.; and Lin, Y.
\newblock 2017.
\newblock Localizing by describing: Attribute-guided attention localization for
  fine-grained recognition.
\newblock In {\em AAAI},  4190--4196.

\bibitem[\protect\citeauthoryear{Liu \bgroup et al\mbox.\egroup
  }{2018}]{retrieval_attentive}
Liu, M.; Wang, X.; Nie, L.; He, X.; Chen, B.; and Chua, T.-S.
\newblock 2018.
\newblock Attentive moment retrieval in videos.
\newblock In {\em SIGIR},  15--24.

\bibitem[\protect\citeauthoryear{Mnih \bgroup et al\mbox.\egroup
  }{2014}]{mnih2014recurrent}
Mnih, V.; Heess, N.; Graves, A.; et~al.
\newblock 2014.
\newblock Recurrent models of visual attention.
\newblock In {\em NIPS},  2204--2212.

\bibitem[\protect\citeauthoryear{Pan \bgroup et al\mbox.\egroup
  }{2016}]{pan2016jointly}
Pan, Y.; Mei, T.; Yao, T.; Li, H.; and Rui, Y.
\newblock 2016.
\newblock Jointly modeling embedding and translation to bridge video and
  language.
\newblock In {\em CVPR},  4594--4602.

\bibitem[\protect\citeauthoryear{Pirinen and
  Sminchisescu}{2018}]{Pirinen_2018_CVPR}
Pirinen, A., and Sminchisescu, C.
\newblock 2018.
\newblock Deep reinforcement learning of region proposal networks for object
  detection.
\newblock In {\em CVPR},  6945--6954.

\bibitem[\protect\citeauthoryear{Plummer \bgroup et al\mbox.\egroup
  }{2015}]{plummer2015flickr30k}
Plummer, B.~A.; Wang, L.; Cervantes, C.~M.; Caicedo, J.~C.; Hockenmaier, J.;
  and Lazebnik, S.
\newblock 2015.
\newblock Flickr30k entities: Collecting region-to-phrase correspondences for
  richer image-to-sentence models.
\newblock In {\em ICCV},  2641--2649.

\bibitem[\protect\citeauthoryear{Regneri \bgroup et al\mbox.\egroup
  }{2013}]{regneri2013grounding}
Regneri, M.; Rohrbach, M.; Wetzel, D.; Thater, S.; Schiele, B.; and Pinkal, M.
\newblock 2013.
\newblock Grounding action descriptions in videos.
\newblock {\em Transactions of the Association of Computational Linguistics}
  1:25--36.

\bibitem[\protect\citeauthoryear{Rohrbach \bgroup et al\mbox.\egroup
  }{2016}]{rohrbach2016grounding}
Rohrbach, A.; Rohrbach, M.; Hu, R.; Darrell, T.; and Schiele, B.
\newblock 2016.
\newblock Grounding of textual phrases in images by reconstruction.
\newblock In {\em ECCV},  817--834.

\bibitem[\protect\citeauthoryear{Shou \bgroup et al\mbox.\egroup }{2017}]{cdc}
Shou, Z.; Chan, J.; Zareian, A.; Miyazawa, K.; and Chang, S.-F.
\newblock 2017.
\newblock {CDC: Convolutional-De-Convolutional} networks for precise temporal
  action localization in untrimmed videos.
\newblock In {\em CVPR},  5734--5743.

\bibitem[\protect\citeauthoryear{Sigurdsson \bgroup et al\mbox.\egroup
  }{2016}]{charades}
Sigurdsson, G.~A.; Varol, G.; Wang, X.; Farhadi, A.; Laptev, I.; and Gupta, A.
\newblock 2016.
\newblock Hollywood in homes: Crowdsourcing data collection for activity
  understanding.
\newblock In {\em ECCV},  510--526.

\bibitem[\protect\citeauthoryear{Singh \bgroup et al\mbox.\egroup
  }{2016}]{localization_multi_stream}
Singh, B.; Marks, T.~K.; Jones, M.; Tuzel, O.; and Shao, M.
\newblock 2016.
\newblock A multi-stream bi-directional recurrent neural network for
  fine-grained action detection.
\newblock In {\em CVPR},  1961--1970.

\bibitem[\protect\citeauthoryear{Sutton and Barto}{2011}]{li2017deep}
Sutton, R.~S., and Barto, A.~G.
\newblock 2011.
\newblock {\em Reinforcement learning: An introduction}.
\newblock Cambridge, MA: MIT Press.

\bibitem[\protect\citeauthoryear{Tellex and Roy}{2009}]{tellex2009towards}
Tellex, S., and Roy, D.
\newblock 2009.
\newblock Towards surveillance video search by natural language query.
\newblock In {\em CIVR},  38:1--38:8.

\bibitem[\protect\citeauthoryear{Tran \bgroup et al\mbox.\egroup }{2015}]{c3d}
Tran, D.; Bourdev, L.; Fergus, R.; Torresani, L.; and Paluri, M.
\newblock 2015.
\newblock Learning spatiotemporal features with {3D} convolutional networks.
\newblock In {\em ICCV},  4489--4497.

\bibitem[\protect\citeauthoryear{Wang \bgroup et al\mbox.\egroup
  }{2018}]{Wang_2018_CVPR}
Wang, X.; Chen, W.; Wu, J.; Wang, Y.-F.; and Yang~Wang, W.
\newblock 2018.
\newblock Video captioning via hierarchical reinforcement learning.
\newblock In {\em CVPR},  4213--4222.

\bibitem[\protect\citeauthoryear{Williams}{1992}]{williams1992simple}
Williams, R.~J.
\newblock 1992.
\newblock Simple statistical gradient-following algorithms for connectionist
  reinforcement learning.
\newblock {\em Machine learning} 8(3-4):229--256.

\bibitem[\protect\citeauthoryear{Xiao, Sigal, and
  Jae~Lee}{2017}]{Xiao_2017_CVPR}
Xiao, F.; Sigal, L.; and Jae~Lee, Y.
\newblock 2017.
\newblock Weakly-supervised visual grounding of phrases with linguistic
  structures.
\newblock In {\em CVPR},  5945--5954.

\bibitem[\protect\citeauthoryear{Xu \bgroup et al\mbox.\egroup
  }{2015a}]{xu2015show}
Xu, K.; Ba, J.; Kiros, R.; Cho, K.; Courville, A.; Salakhudinov, R.; Zemel, R.;
  and Bengio, Y.
\newblock 2015a.
\newblock Show, attend and tell: Neural image caption generation with visual
  attention.
\newblock In {\em ICML},  2048--2057.

\bibitem[\protect\citeauthoryear{Xu \bgroup et al\mbox.\egroup
  }{2015b}]{xu2015jointly}
Xu, R.; Xiong, C.; Chen, W.; and Corso, J.~J.
\newblock 2015b.
\newblock Jointly modeling deep video and compositional text to bridge vision
  and language in a unified framework.
\newblock In {\em AAAI},  2346--2352.

\bibitem[\protect\citeauthoryear{Yeung \bgroup et al\mbox.\egroup
  }{2016}]{lifeifei_loc}
Yeung, S.; Russakovsky, O.; Mori, G.; and Fei-Fei, L.
\newblock 2016.
\newblock End-to-end learning of action detection from frame glimpses in
  videos.
\newblock In {\em CVPR},  2678--2687.

\bibitem[\protect\citeauthoryear{Zhao \bgroup et al\mbox.\egroup }{2017}]{ssn}
Zhao, Y.; Xiong, Y.; Wang, L.; Wu, Z.; Tang, X.; and Lin, D.
\newblock 2017.
\newblock Temporal action detection with structured segment networks.
\newblock In {\em ICCV},  2914--2923.

\end{thebibliography}

\end{document}